%% file: main.tex
\documentclass[conference,a4paper]{IEEEtran}
\usepackage{graphicx}
\usepackage{cite}
\usepackage{amsmath,amssymb,amsfonts}
\usepackage{algorithmic}
\usepackage{textcomp}
\usepackage{wrapfig}

\usepackage{cleveref}
\usepackage{multirow}

\def\BibTeX{{\rm B\kern-.05em{\sc i\kern-.025em b}\kern-.08em
    T\kern-.1667em\lower.7ex\hbox{E}\kern-.125emX}}

\begin{document}
\title{MultiBARF: Integrating Imagery of Different Wavelength Regions by Using Neural Radiance Fields}
\author{\IEEEauthorblockN{Kana Kurata, Hitoshi Niigaki, Xiaojun Wu, and Ryuichi Tanida
\IEEEauthorblockA{NTT Corporation, Japan}}
}

\maketitle

\input{sec/0_abstract}
\includegraphics[width=2.9in]{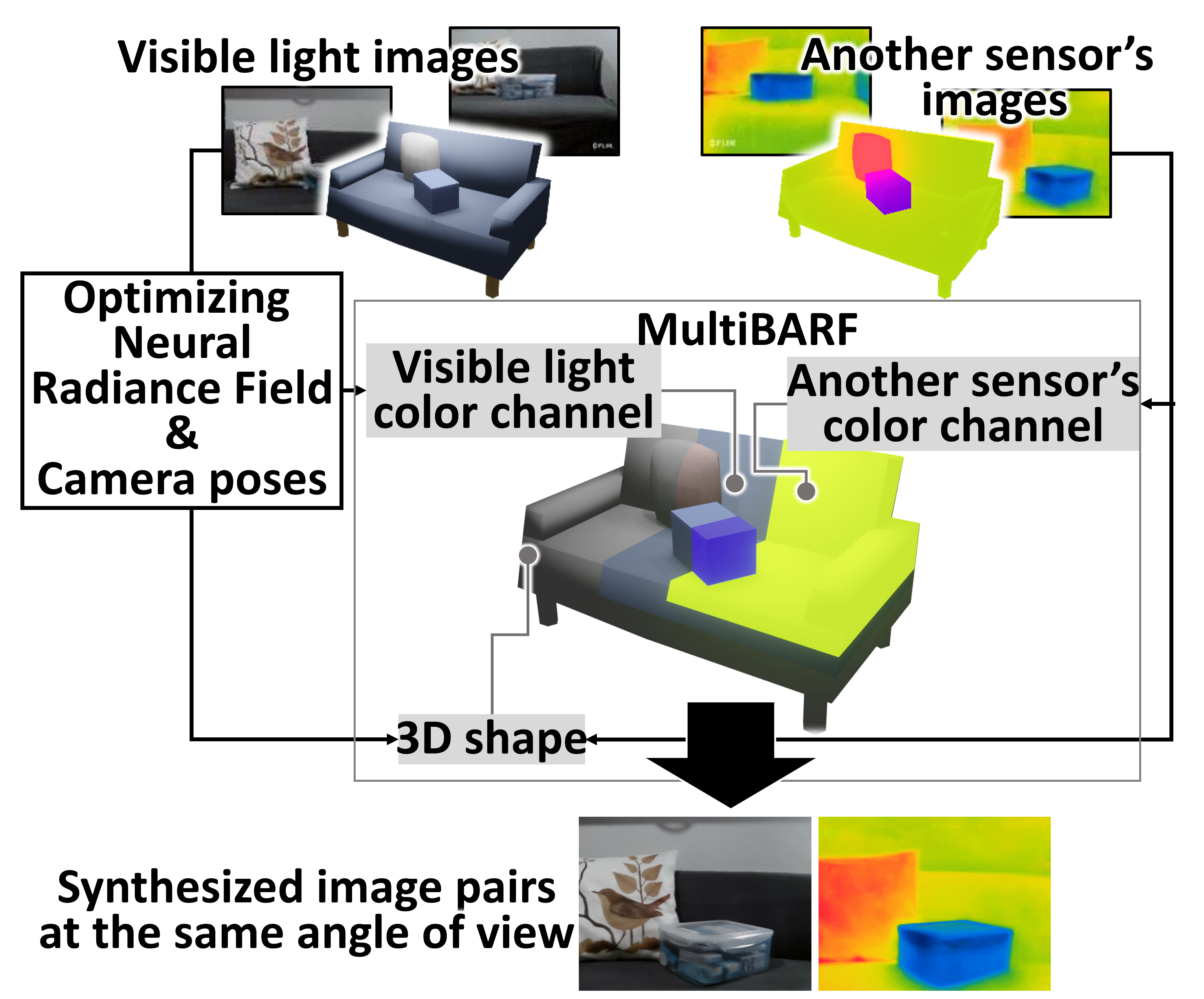}%

\input{sec/1_intro}

\input{sec/2_related_work}

\input{sec/3_approach}
\input{sec/4_experiments}
\input{sec/5_conclusion}

\bibliographystyle{IEEEtran}
\bibliography{main}

\end{document}

%% file: sec/0_abstract.tex
\begin{abstract}
Optical sensor applications have become popular through digital transformation. Linking observed data to real-world locations and combining different image sensors is essential to make the applications practical and efficient. However, data preparation to try different sensor combinations requires high sensing and image processing expertise. To make data preparation easier for users unfamiliar with sensing and image processing, we have developed MultiBARF. This method replaces the co-registration and geometric calibration by synthesizing pairs of two different sensor images and depth images at assigned viewpoints. Our method extends Bundle Adjusting Neural Radiance Fields(BARF), a deep neural network-based novel view synthesis method, for the two imagers. Through experiments on visible light and thermographic images, we demonstrate that our method superimposes two color channels of those sensor images on NeRF.
\end{abstract}

%% file: sec/1_intro.tex
\section{Introduction}\label{sec:intro}

\IEEEPARstart{S}{ensor} applications,  consisting of sensing, data processing, and visualization or machine automation techniques, have become standardly used daily through digital transformation. Optical imagers are especially popular for recognizing the environments around us: visible light cameras for driving assistance, near-infrared cameras for non-contact pill inspection, and thermal infrared cameras for body temperature measurement at building entrances. These applications often link observed data to real-world locations for display to humans and automatic machine control. Also, combining multiple different sensors is a popular technique~\cite{rs-fusion-review}. For example, combinations of optical imagers of different wavelength regions are effective remote sensing for environmental monitoring\cite{EnvironmentalRemotesensingReview}. For another example, driving assist systems use visible light cameras combined with infrared cameras, also, according to the monitoring target or the surrounding conditions, with active sensors such as radars and LiDARs\cite{electronics11142162,s21165397}. These examples show the importance of selecting appropriate sensor combinations for targets and observation environments.

% The popularization of sensor applications is supported by utilities that do not require user expertise. But still, establishing the core algorithm of an application for a specific purpose is complicated, even by using machine learning. Because it requires expertise in all analysis targets, sensing, and image processing. On the contrary, experts in particular fields sometimes acquire much information just by reading low-level processed sensor images. This step, imagery interpretations by experts of analysis targets, is essential to create new sensor applications for specific cases.

Data acquisition and Preprocesses are necessary tasks for sensor applications. Preprocesses have to make the dataset comprehensively analyzable by post-processes of applications. Sometimes, dataset providers distribute co-registered products of multiple sensor images like ~\cite{Landsat8}. Also, providing applications bounded with the equipment sets, like driving support systems, decreases users' efforts in data preparation. Such simplifications in data preparation support the recent popularization of sensor applications. 

% Sometimes, multiple sensor images with registration are provided in remote sensing, like ~\cite{Landsat8}.
% in sensing, image processing, or machine learning (depending on the case)

On the other hand, trying several sensor combinations without the limitation of existing equipment combo still requires high sensing and image processing skills. To prepare custom datasets consisting of multiple sensor imagery and linked with location information, the preprocess must include radiometric/geometric corrections for each sensor image, co-registration between different sensors, and geometric calibration of all dataset images. While radiometric/geometric corrections depend on sensor properties, co-registration and geometric calibration are often communalized for different types of sensors. We have therefore focused on making the co-registration and geometric calibration processes not require high expertise and be adaptive to various combinations of sensors.

% geometric camera calibration to define the position of images on a coordinate system (e.g., relative locations among the images or, on local 3D models, absolute locations on a geographic coordinate system). 

 % There are tips on the web for image processing techniques for visible light cameras, but not for other sensors. Also, the methods for visible light images often cannot be applied to other kinds of images.
 % For example, camera resectioning techniques for visible light images can not used for images of certain wavelength regions because those images don't have enough texture for feature detection.

% combining multiple sensors and linking them to 3D locational information, and approaching the problems of pixel-wise registration and 3D reconstruction.

% This paper aims to support the development of multi-sensor applications in appropriate sensor combinations by providing a way to easily prepare multi-sensor datasets. 

This paper proposes MultiBARF, a method to synthesize pairs of two different sensor images and depth images (2D rendered 3D shape) at assigned viewpoints by only inputting multiview images of two sensors into the model. Neural Radiance Fields (NeRF)\cite{nerf} is a successful deep learning-based 3D representation for photorealistic scenes. NeRF records 3D models as color and density distributions by a Deep Neural Network(DNN). It is optimized with multiview images and visualized through volume rendering. MultiBARF is based on BARF\cite{barf}, an expansion of NeRF having a function of bundle adjustment. We enabled BARF to work for two different sensor images and automatically register them. Our method does not directly solve the co-registration and geometric calibration but can replace these processes.

The characteristics of our method are that it extends the DNN structure of NeRF to be used for two different imagers and does not require camera calibration between the two imagers by utilizing the camera pose estimation process of Bundle Adjusting Neural Radiance Fields(BARF)\cite{barf}, a kind of NeRF, for the two imagers.

We evaluated our model using a combination of visible light and thermal infrared images since thermal infrared images are a challenging target for 3D reconstruction and registration in daily scenes. We demonstrate that MultiBARF synthesizes an image pair of two image sensors, observing different wavelength regions at a single viewpoint.

In summary, our contributions are:
\begin{itemize}
  \item We propose MultiBARF, a new architecture of NeRF that has two color channels for two different image sensors. 
  \item We provide a method to optimize MultiBARF depending on two sensor images without complicated camera calibration.
  \item We demonstrate the synthesizing results of MultiBARF on visible light and thermographic images.
\end{itemize}

%% file: sec/2_related_work.tex
\section{Related Work}\label{sec:related_work}
This study approaches the sensing problem of preparing multispectral datasets using a computer vision technique called NeRF\@. Therefore, we introduce related methods from the following perspectives: applications in multiple wavelength regions, preparation of multispectral datasets, and NeRF with its extensions.

\leavevmode \\
\textbf{Optical imager applications.}
There are many applications in the visible and infrared regions, as these areas contain abundant information and are observable in non-special environments. As an object's condition appears as a reflection or emission feature at the corresponding wavelength, choosing the optimal observation wavelengths is essential.

Ultraviolet (UV), visible (VIS), and near-infrared (NIR) to short-wavelength infrared (SWIR) mainly relate to chemical composition or surface microstructure. The UV region can watch the characteristics relating to the attractiveness of a flower to insects~\cite{WU2023}. Conditions of vegetations mainly appear in NIR ~\cite{KNIPLING1970155}, but combined with VIS and SWIR~\cite{LI2021112230} improved the estimation accuracy for leaf water content. For another example, combining VIS to SWIR~\cite{geological_mapping} discriminates minerals that seem to be just whitish rocks. 

On the other hand, the long-wavelength regions called long-wavelength infrared (LWIR) or thermal infrared (TIR), relating to the temperature of daily scenes, sometimes show objects' internal conditions. Facilities' internal deteriorations make temperature differences on wall surfaces over them and captured by can be revealed by thermography (TIR region)~\cite{ndt_ir, s22020423}. For another example, Maierhofer et al.~\cite{MAIERHOFER2006393} estimate internal void and honeycombing of concrete.

The above examples show that multispectral datasets can help us acquire independent information from different wavelength regions and estimate the targets' comprehensive conditions.

\leavevmode \\
\textbf{Preparation of multispectral datasets.}
Co-registration between different sensors is necessary to use multiple wavelength regions because a single imager supports a limited wavelength range. As previously reviewed, many effective methods have been developed for datasets with different characteristics~\cite{survey_registration_1992, ZITOVA2003977}. 

Also, linking locational information (e.g., relative locations among the dataset or on 3D models, absolute locations on GIS) through geometric camera calibration is essential to identify the analyzed targets in real space. For the VIS region, there are many convenient calibration tools like COLMAP(\cite{schoenberger2016sfm},\cite{schoenberger2016mvs}), generally working near VIS (UV, NIR, SWIR). However, these methods require hardware-level customization for long-wavelength regions~\cite{thermal-calib}. Although we can measure camera parameters alternatively when shooting, it requires additional equipment and expertise.

\leavevmode \\
\textbf{Neural Radiance Fields (NeRF).}
Mildenhall et al.~\cite{nerf} proposed the novel view synthesis method that represents photorealistic continuous scenes by ``Neural Radiance Fields'' (NeRF). Unlike conventional 3D representations such as voxel or mesh, NeRF expresses 3D scenes or objects by continuous functions using Deep Neural Networks (DNNs). It expresses 3D scenes as the distribution of density and color and is rendered into images by classic volume rendering techniques. 

There are many extensions of NeRF, but we have yet to find previous research targeting the same situation as this paper. Thus, we mention some related works from two aspects: geometric camera calibration and modeling multiple appearances for a scene. 

For the former, we have two options: calibration processes are coupled with the optimizing process of NeRF tightly or loosely. NeRF techniques require the intrinsic and extrinsic camera parameters for training images. Loosely coupled methods can use hardware or software calibration methods. On the other hand, \textit{Ne{RF}-R-}~\cite{nerf-mm} and BARF~\cite{barf} estimate camera parameters along with optimizing NeRF. These methods give an initial camera pose for every training image, then optimize simultaneously to train NeRF. Ne{RF}$--$~\cite{nerf-mm} also estimates intrinsic camera parameters.

For the latter, NeRF in the Wild~\cite{nerf-iw} is a representative technique. This technique can handle scenes with different sunlight and lighting in each image. It can learn the static and changing parts of scenes as different parameters and express lighting changes as an appearance embedding vector for each image. Furthermore, Ha-NeRF ~\cite{ha-nerf} expands the NeRF in the Wild to generate appearance embedding from any images. This method uses a convolutional neural network (CNN) based appearance encoder. Thus, the CNN trained with a dataset can convert any images into vectors, even those not used for training.

%% file: sec/3_approach.tex
\section{Approach}\label{sec:approach}
Our MultiBARF replaces co-registration and geometric calibration processes with a simple method for synthesizing superimposed pair images of two different image sensors with location on a reconstructed local 3D model. The technique is an extension of NeRF that can represent two registered sensor color channels. This section describes the DNN structure and functions of MultiBARF and its optimization process.

\subsection{Structure of MultiBARF}
Our method derives from NeRF and BARF. We first mention these base methods, followed by the differences between our model and them in structure and function.

\begin{figure}[b]
  \centering
    \includegraphics[width=1.0\linewidth]{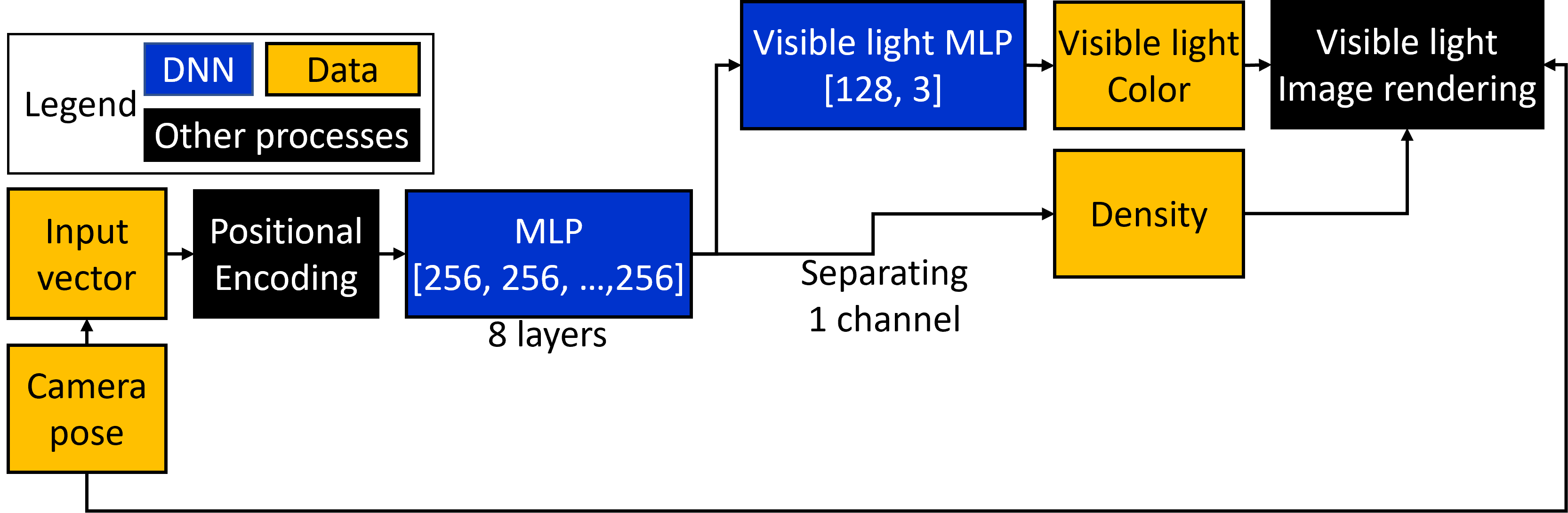}
  \caption{The original BARF structure.}\label{fig:DNN_BARF}
\end{figure}

\leavevmode \\
\textbf{NeRF and BARF.} 
NeRF~\cite{nerf} is a novel view synthesis method that represents photorealistic scenes by following steps. First, it queries vectors storing spatial location and viewing direction along camera rays. Next, a continuous function converts the vectors into color and density for the queried locations. Lastly, those colors and density at queried locations are composited into the images by classic volume rendering techniques. The continuous function of the second step is a composite function of untrainable positional encoding and following trainable DNN. The positional encoding maps the inputs to a higher-dimensional space using high-frequency functions. The DNN consists of fully connected (non-convolutional) layers and is optimized based on training photos with known camera poses to make the composited images at those camera poses get closer to the training images. Through the above processes, the model reconstructs a colored 3D model as the density and color distribution. Still, sometimes, the object's surface may not be correctly reconstructed depending on the narrowness of the learned angle of view.

BARF~\cite{barf} follows the structure of NeRF as shown in (\cref{fig:DNN_BARF}) but additionally has the capability of registration. Its registration method theoretically derives from classical image alignment and adopts a coarse-to-fine strategy. At first, every training image has an initial camera pose. BARF optimizes the pose by calculating the appropriate viewpoint warp to minimize the photometric error between training and synthesized images. This process works after every single epoch of DNN training. The coarse-to-fine strategy is included in positional encoding. It controls the DNN to learn low-frequency features at first and, lately, covering high-frequencies to prevent the positional encoding of NeRF harms the camera pose optimization. We refer the reader to the original papers \cite{nerf, barf} to get descriptions by illustrations or formulas.

\begin{figure}[b]
  \centering
    \includegraphics[width=1.0\linewidth]{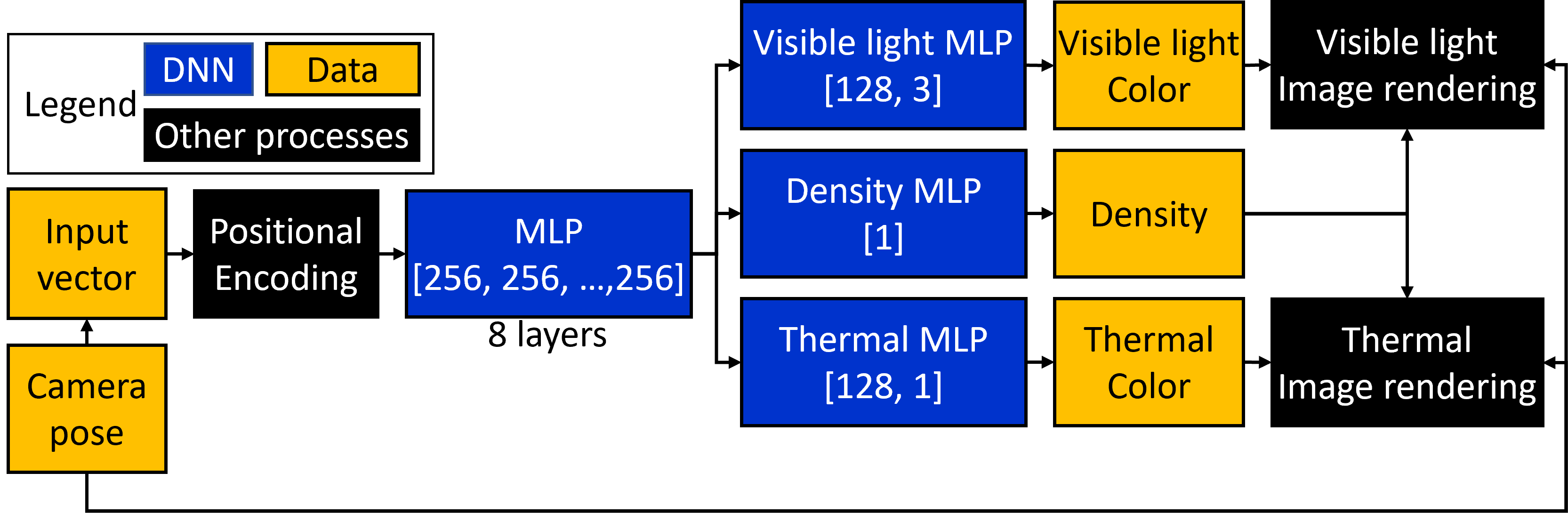}
  \caption{Our MultiBARF structure.}\label{fig:DNN_ours}
\end{figure}

\leavevmode \\
\textbf{MultiBARF.} 
Our model's key structure is the two sensor color channels and a density channel, while the original NeRF/BARF has one color channel and one density channel (\cref{fig:DNN_ours}). Here, we present our model with an example that combines the visible and thermal infrared wavelength ranges. Still, other combinations are available as long as both sensors can capture identical objects and one sensor includes enough features for 3D reconstruction. The post-branched two MLPs, "visible light MLP" and "Thermal MLP," convert the output of thick pre-branched MLP into colors of visible light and thermal infrared images. 

On the other hand, MultiBARF has just one density channel and uses this same density distribution to synthesize both sensor images. This density-sharing structure is the key to approaching registration and 3D reconstruction simultaneously. Our approach assumes that objects' 3D shapes are immutable, even between different sensors. The spacial gaps based on 3D shapes appear in images as color boundaries. Even if the textures or surface paintings appear not to be shared between different sensor images, the boundaries derived from 3D shapes are common. 
Our model's density-sharing structure is designed to be consistent with this situation. The density distribution works as the standard of coordinate systems between two different sensor images without known camera poses. To make this structure work, the model always uses pre-branched and density MLPs to synthesize any sensor's images while switching color-output MLPs based on the target sensor.

From the registration aspect, this density-sharing structure makes the camera pose optimization process work for two different sensor images based on the same geometries stored in the density. From 3D reconstruction, this structure can learn 3D geometries with texture-less sensor images like thermal infrared images by supplementing the 3D shape information from images that include abundant textures like visible light, ultraviolet, or near-infrared. About the textures, the structure that color-output MLPs are not shared allows the model to store different textures efficiently by only sorting out the color boundaries of each sensor's images to the related color-output sub-MLPs\@.

\begin{figure}[tb]
  \centering
    \includegraphics[width=0.45\linewidth]{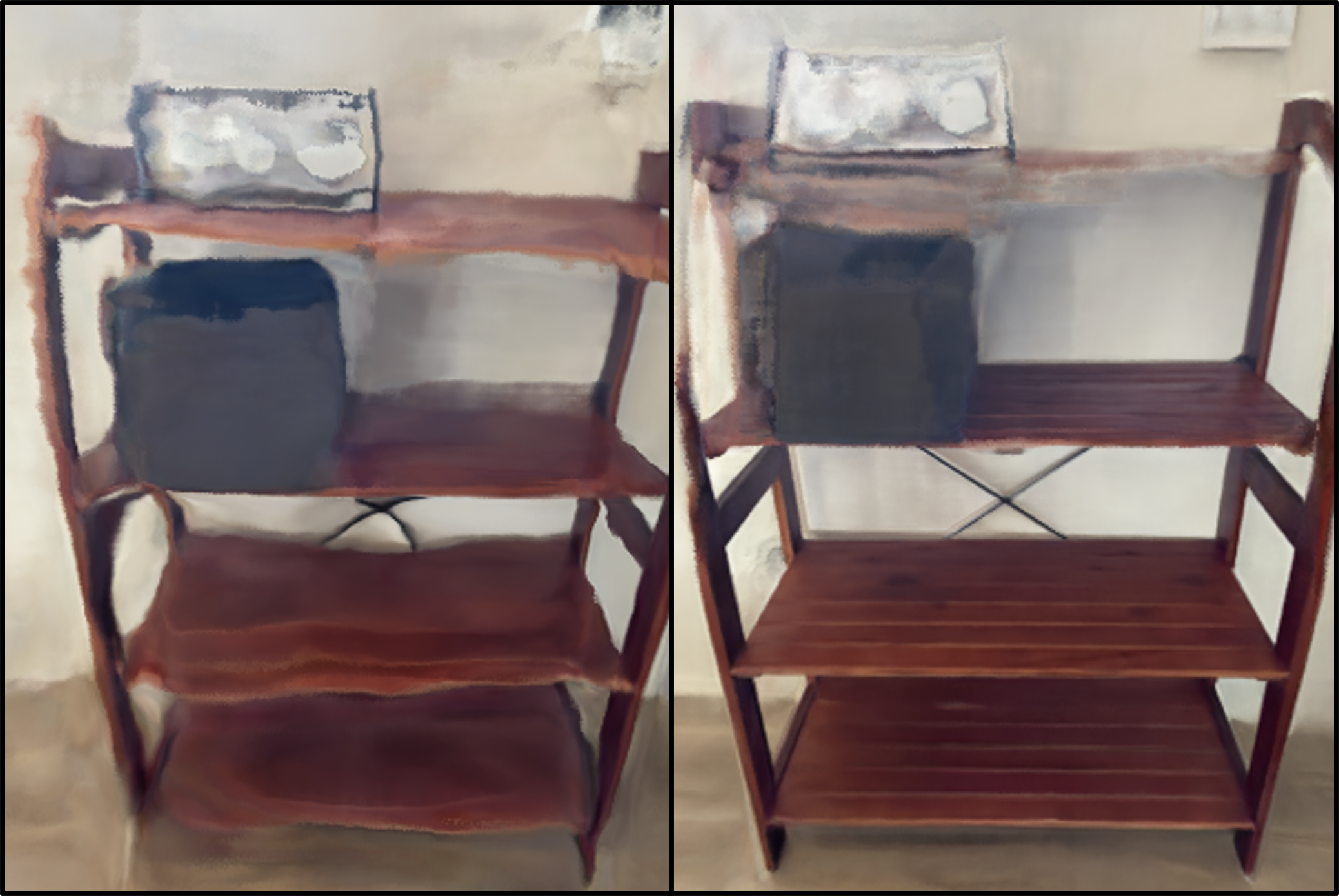}
  \caption{The results with various balances of pre and post-branch layers. From the left, the number of layers of pre and post-branch are 4--6 and 8--2.}\label{fig:res_layer}
\end{figure}

Also, we mention detailed network architecture here. We use post-branched MLPs with a much smaller number of layers than pre-branched. This balance is according to an experiment. The original BARF model has a branch of density and color. We experimented with changing the balance of the number of layers before and after the branch while keeping the whole number of layers. \cref{fig:res_layer} shows the results of patterns in which the number of pre and post-branch layers are 4--6 and 8--2. The shorter we set the pre-branch, the more the object shape degraded. Therefore, our model has thick pre-branch layers and thin post-branch MLPs.

\subsection{Optimization of MultiBARF}
MultiBARF has optimization modes for visible light images and thermal images. Each mode's process follows the BARF training method to make the model's functions derived from BARF work appropriately. However, each mode switches the modules to be optimized.

When we train the model with visible light images, the DNN is optimized depending on the loss calculated between the synthesized and training visible light images. Then, camera poses of visible light training images are optimized as BARF\@. This phase does not use thermal color-output MLP\@. Therefore, the training losses are back-propagated to MLPs other than the thermal color-output MLP\@. The visible light and thermal color-output MLPs are optimized with each sensor's image, while the density and the pre-branch MLPs are optimized with both sensor images. For example, the model does not train visible light color output-MLP in the case of training with thermal images. 

Then, there is an option to learn the two sensor images in any order. Let us jump to the point: In this paper, we chose the way that uses the two sensors alternatively. This method matches the coarse-to-fine approach of BARF. In addition to the advantages mentioned in the original BARF, we have another reason to focus on the coarse-to-fine process. In the general scene, higher-frequency features tend to relate to visible light textures. It means that lower-frequency features relate more to object shapes and are likely to appear more commonly between different sensors. Thus, we adopt a harmonious approach with the coarse-to-fine process to handle multiple sensor images.
\cref{sec:exp_train} describes the detailed comparison of multiple optimization methods.

%% file: sec/4_experiments.tex
\section{Experiments}\label{sec:experiments}
This section evaluates the performance of MultiBARF and its training method. First, we assessed the basic ability and practical performance of our model. \cref{sec:exp_restricted} discusses the basic ability of the model by evaluating with restricted datasets that consist of almost the same camera pose pairs of visible light and thermal images. In contrast, \cref{sec:exp_practical} evaluates the practical performance on datasets with no correspondence between camera poses of visible light and thermal images. The last \cref{sec:exp_train} compares three types of training order and confirms that the appropriate training method is alternatively learning visible light and thermal images.

\begin{figure}[t]
    \centering
      \includegraphics[width=1.0\linewidth]{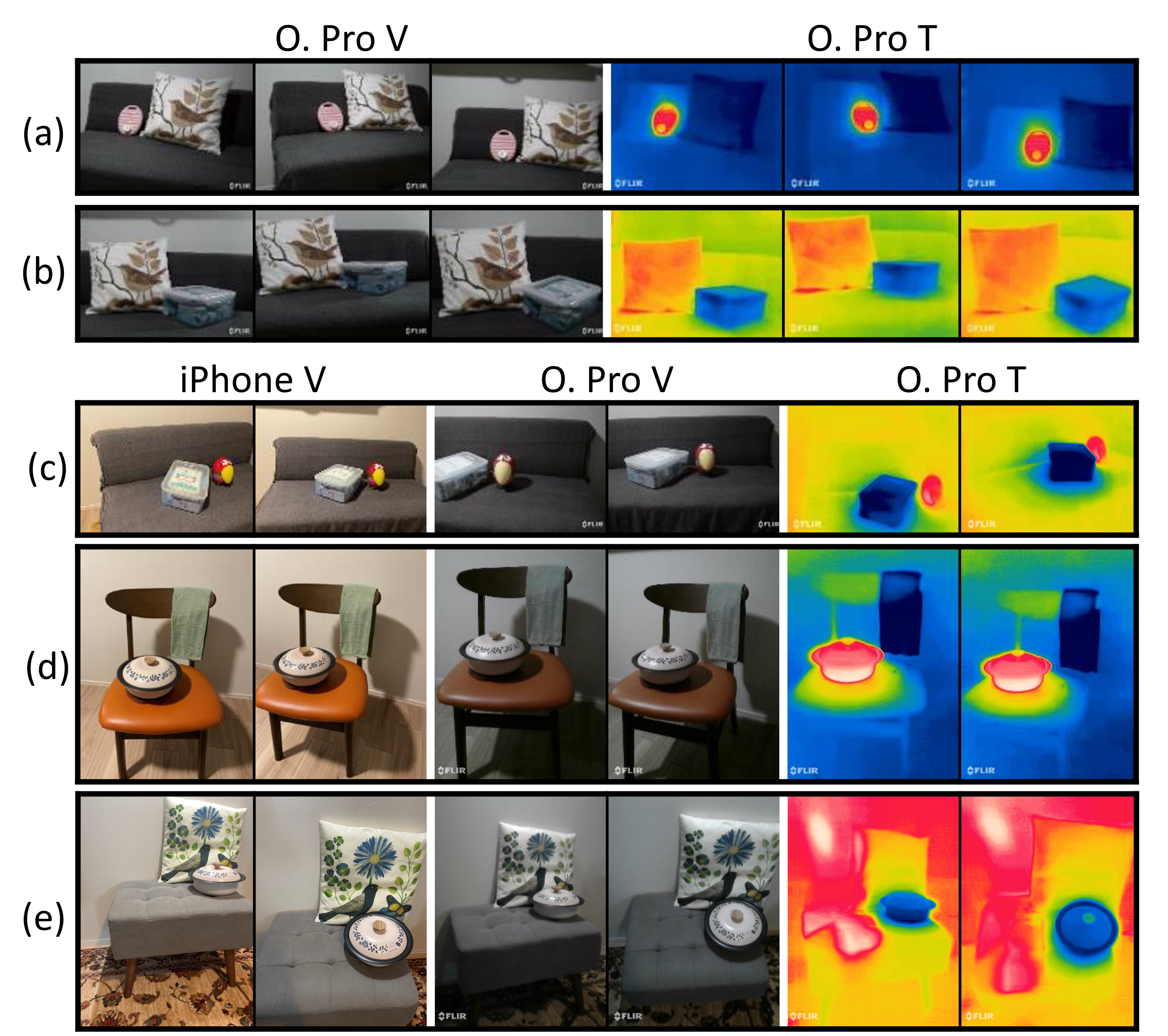}
    \caption{A part of the datasets; (a)``Warmer'', (b)``Pack'', (c)``Toy'', (d)``Casserole'', and (e)``Trace''.}
    \label{fig:datasets}
\end{figure}

\renewcommand{\arraystretch}{1.1}
\begin{table}[b]
  \caption{The number of images of each camera in the datasets.}
  \centering
  \setlength{\tabcolsep}{2.0pt}
  \begin{tabular}{l|c|c|c}
      \hline\noalign{\smallskip}
         \multirow{2}{*}{Scene} & \multicolumn{3}{c}{Qty. of}  \\                     
                          & iPhone V & O.Pro V & O.Pro T \\
         \hline
                            Warmer & - &16 & 16 \\
                            Pack & - & 22 & 22 \\
                            Toy & 16 & 16 & 16 \\
                      Casserole & 21 & 27 & 27 \\
                      Trace & 20 & 20 & 20 \\
         \hline
  \end{tabular}
  \label{tab:qty_images}
\end{table}

We created a new dataset since no publicly available dataset exists for testing view synthesis or multiview 3D reconstruction techniques consisting of visible and invisible light images. Equipment and experimental settings are shared in \cref{sec:exp_restricted} and \cref{sec:exp_practical}.

\leavevmode \\
\textbf{Equipments.} We used the iPhone 12 Pro Max and FLIR ONE Pro for image acquisition. Our method only requires the internal camera parameters to use these devices as BARF. Generally, we can obtain the internal parameters from the device specifications mentioned below. 
 
The iPhone 12 Pro Max has Ultra Wide, Wide, and Telephoto cameras and LiDAR. In this study, we only used the Wide camera. Its aperture is ƒ/1.6, corresponding to the focal length of 26mm. With FLIR ONE Pro, we shoot images with a 4:3 aspect ratio.
 
FLIR ONE Pro is an iPhone accessory camera. In this experiment, we used it connecting to the iPhone 12 Pro Max. Invisible light image sensors like this have become more popular and readily available. FLIR ONE Pro has visible light and thermal cameras 10 mm apart to capture a pair of visible light and thermal photos with a slight positional offset in a single shot. The horizontal and vertical camera angles are 55°± 1°and 43°± 1°with an aspect ratio of 4:3 for both visible light and thermal cameras. The resolution is 1440$\times$1080 for visible light images and 160$\times$120 for thermal images. The appearance of thermal images changes quickly by heat conduction when temperature distribution is biased, which limits the number of shots because we must acquire a dataset within which the temperature distribution does not change significantly (within about 2 minutes in this study).
 
For efficiency and meeting conditions among the cameras, We resized all images to a width and height of 400$\times$300 pixels (or transpose for a vertically elongated dataset).

\leavevmode \\
\textbf{Common experimental conditions.} In all the experiments of this paper, we did not explicitly input information about the positional relationships, and the camera positions were estimated individually. The hyperparameters of BARF follow the experiment of BARF on the LLFF dataset (\cite{llff}) in Lin et al.~\cite{barf}. Training batches of visible light and thermal mode consist of 4096 pixels randomly chosen from all corresponding training images. Every FLIR ONE Pro image has a logo, so we excluded the pixels of the logo from the training batch selection. For evaluation,  We split 13\% of (rounding down to integer) images for validation and the rest for training. In this paper, we present the results after training 200000 epochs.

\leavevmode \\
\textbf{Abbreviation in figures and tables.}
In figures and tables below, ``FLIR ONE Pro'' is shortened to ``O.Pro''. Also, ``visible light image'' and ``thermal image'' are stated as ``V'' and ``T'' respectively.

\renewcommand{\arraystretch}{1.1}
\begin{table}[b]
  \caption{Quantitative results (Mean PSNR, SSIM, and LPIPS) for \textbf{visible light images} of \textbf{training data} in basic performance evaluation.}
  \centering
  \setlength{\tabcolsep}{2.0pt}
  \begin{tabular}{c|l|c|c|c}
    \hline\noalign{\smallskip}
       Scene & Method & PSNR ↑ & SSIM ↑ & LPIPS ↓ \\
    \hline
        \multirow{2}{*}{Warmer} & Ours & \textbf{17.94} & \textbf{0.66} & \textbf{0.32} \\ 
                                & BARF w/ O.Pro V  & 14.25 & 0.49 & 0.52 \\
    \hline
        \multirow{2}{*}{Pack} & Ours               & 16.84 & 0.60 & 0.42 \\ 
                              & BARF w/ O.Pro V    & \textbf{19.69} & \textbf{0.69} & \textbf{0.27} \\
    \hline
  \end{tabular}
  \label{tab:result_1}
\end{table}

\subsection{basic performance}\label{sec:exp_restricted}

We evaluate MultiBARF's essential ability to learn and register the color distributions of two different sensors. We designed this experiment to confirm the following three perspectives.
 
First is the 3D shape reconstruction. We have to confirm that MultiBARF improves the result of density distribution optimized with both sensors' images compared to BARF with thermal images. Next is the registration of learned color distributions for two sensors. They should be consistent with the density distribution learned with both sensors. The last is the separation of the image characteristics, which means the synthesized images for each sensor must only correspond to each sensor's pictures and not contain attributes of the other sensor images. We evaluated these issues qualitatively. Also, we assessed the quantitative degradation of the visible light results compared to the original BARF learning only visible light images.

\leavevmode \\
\textbf{Datasets.} We prepared datasets, ``Warmer'' and ``Pack,'' containing FLIR ONE Pro's visible light and thermal images. These datasets consist of visible light and thermal photo pairs. ``Warmer'' captures a scene of a cushion and a hot water bottle (\cref{fig:datasets} (a)) on a sofa, and ``Pack'' consists of a transparent case filled with sundry goods and a cushion (\cref{fig:datasets} (b)) on a sofa. The number of images for each dataset is as in \cref{tab:qty_images}.

\renewcommand{\arraystretch}{1.1}
\begin{table}[b]
  \caption{Quantitative results for \textbf{visible light images} of \textbf{validation data} in basic performance evaluation.}
  \centering
  \setlength{\tabcolsep}{2.0pt}
  \begin{tabular}{c|l|c|c|c}
    \hline\noalign{\smallskip}
       Scene & Method & PSNR ↑ & SSIM ↑ & LPIPS ↓ \\
    \hline
        \multirow{2}{*}{Warmer} & Ours & \textbf{14.32} &\textbf{ 0.48} & 0.58 \\ 
                                & BARF w/ O.Pro V  & 13.57 & 0.46 & \textbf{0.56} \\
    \hline
        \multirow{2}{*}{Pack} & Ours               & 14.22 & 0.52 & 0.53 \\ 
                              & BARF w/ O.Pro V    & \textbf{18.65} & \textbf{0.68} & \textbf{0.24} \\
    \hline
  \end{tabular}
  \label{tab:result_1-val}
\end{table}
\begin{figure}[t]
  \centering
    \includegraphics[width=0.85\linewidth]{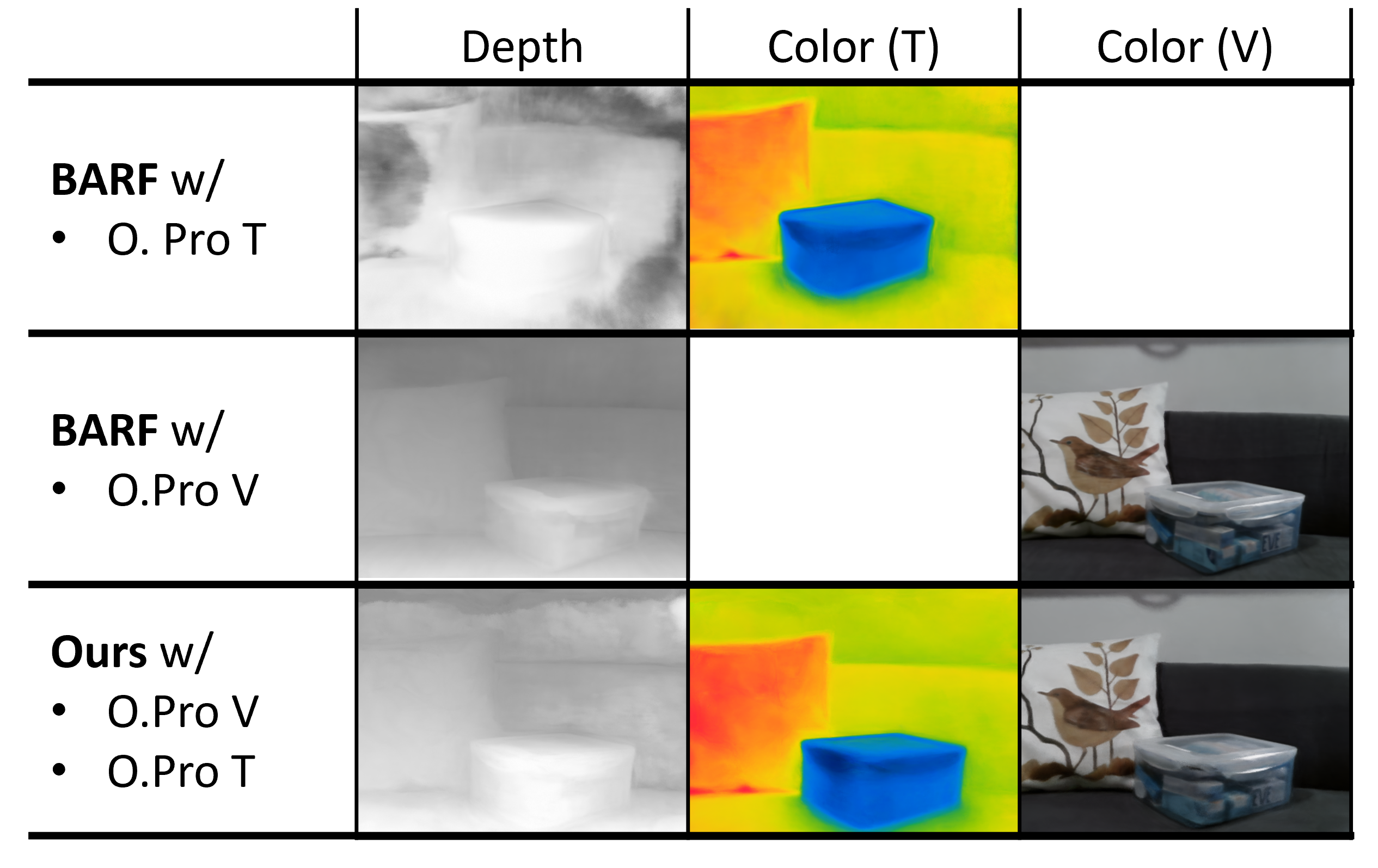}
  \caption{The synthesized results of the ``Pack'' dataset. The top and middle rows show the synthesized depth and color images by BARF trained with only visible light or thermal images, respectively. The bottom is the results of our MultiBARF, which learned both visible light and thermal images.}
  \label{fig:result_1}
\end{figure}
\leavevmode \\
\textbf{Results.} 
First, we discuss the term 3D shape reconstruction. See \cref{fig:result_1}. The top and middle rows show BARF results optimized with each sensor's images, and the bottom is MultiBARF with both sensor images. In the top row, the 3D geometries were not reconstructed consistently with the actual object shapes by BARF with only thermal images. In contrast, the depth of our model shows the shape of the sofa, cushion, and pack so that it is successfully optimized along with the object shapes. It suggests our model improves 3D reconstruction results with thermal images by supplementing information from visible light images.

To discuss registering learned color distributions, focus on the lower row of \cref{fig:result_1}. A part of the color changes appears in the lower center image, the thermal image output. Their locations are consistent with boundaries formed in the depth image on the left. However, some boundaries, such as between the backrest and the seating face of the sofa, are not shown in the thermal result. This characteristic is because the temperature and emissivity do not change between them, the same as the training data of \cref{fig:datasets} (a), which is the correct result for thermal images. In the synthesized visible light image on the right, the boundaries, including no color changes in thermal images, appeared at the location consistent with the depth image. These results show that our model can register the color distribution of two sensors.
 
The latter part of the above discussion relates to separating the image characteristics. The unique patterns of visible light images do not appear in the synthesized thermal image but appear correctly in the visible light result. Unlike thermal images, the depth image exhaustively contains boundaries corresponding to geometries but does not contain visible light textures. 

Also, we mention the quality degradation of the visible light results. The quantitative results (PSNR, SSIM, and LPIPS, the same metrics as the original NeRF and BARF) for training and validation data (\cref{tab:result_1,tab:result_1-val}) show that there is a tendency based on the scenes but no systematic degradation. These results indicate that our method successfully reconstructed the scene characteristics of textures and 3D geometries from visible light images even when learning additional sensor images.

From the above discussions, we confirmed that our approach could match the coordinate systems of density and two sensor's color channels and express the boundaries of colors not shared by all sensors.

\renewcommand{\arraystretch}{1.1}
\begin{table}[b]
  \caption{Quantitative results (mean PSNR, SSIM, and LPIPS) for \textbf{visible light images} of \textbf{training data}.}
  \centering
  \setlength{\tabcolsep}{2.0pt}
  \begin{tabular}{c|l|c|c|c}
    \hline\noalign{\smallskip}
       Scene & Method & PSNR ↑ & SSIM ↑ & LPIPS ↓ \\
        \hline
        
              \multirow{4}{*}{Toy} & Ours w/ O.Pro V & 14.02 & 0.51 & 0.65 \\
                                   & BARF w/ O.Pro V & \textbf{19.63} & \textbf{0.70} & \textbf{0.41}\\
                \cline{2-5}
                & Ours w/ iPhone V & 14.06 & 0.50 & 0.58 \\
                & BARF w/ iPhone V & \textbf{16.68} & \textbf{0.53} & \textbf{ 0.41} \\
              \hline
              
              \multirow{4}{*}{Casserole} & Ours w/ O.Pro V & 18.75 & 0.64 & 0.41 \\
                                         & BARF w/ O.Pro V & \textbf{ 20.36} & \textbf{0.68} & \textbf{0.35} \\
                  \cline{2-5}
                & Ours w/ iPhone V &\textbf{ 16.42} &\textbf{ 0.60} & \textbf{0.40} \\
                & BARF w/ iPhone V & 14.92 & 0.57 & 0.42 \\
              \hline

              \multirow{4}{*}{Trace} & Ours w/ O.Pro V & 15.31 & 0.46 & 0.59 \\
                                         & BARF w/ O.Pro V & \textbf{19.27} & \textbf{0.60} & \textbf{0.44} \\
                  \cline{2-5}
                & Ours w/ iPhone V &\textbf{14.30} &\textbf{0.38} & \textbf{0.61} \\
                & BARF w/ iPhone V & 14.50 & 0.40 & 0.55 \\
              \hline

    \hline
  \end{tabular}
  \label{tab:result_2}
\end{table}
\setlength{\tabcolsep}{1.4pt}

\subsection{Practical performance}\label{sec:exp_practical}
We evaluate the model on more practical conditions. We have many options for sensor selection. We can use a built-in combination like FLIR ONE Pro or a combination of a smartphone camera and another image sensor. This section tries the condition that iPhone and FLIR ONE Pro images have no camera pose relationship. We compare three patterns: BARF optimized with thermal photos, MultiBARF with visible light and thermal images of FLIR ONE Pro, and MultiBARF with the iPhone's visible light and FLIR ONE Pro thermal images. In the last pattern, the angle of view differs between the iPhone visible light images and FLIR ONE Pro thermal images. 

\leavevmode \\
\textbf{Datasets.} We made the ``Toy'', ``Casserole'' and ``Trace'' datasets (\cref{fig:datasets} (c) - (f)) to evaluate our model's performance using multiple sensors. There are no positional relationships between iPhone and FLIR ONE Pro images. The dataset consists of visible light images taken by the iPhone built-in camera and thermal images taken by FLIR ONE Pro. 
The number of images for each dataset is as in \cref{tab:qty_images}. We arranged the region of shooting positions to be almost the same between FLIR ONE Pro and iPhone, but the difference in focal length between the cameras makes a slight difference in the captured area. Also, to evaluate the effects based on the implicit camera pose relationships, no camera pose relationships exist between the visible light images and thermal images of FLIR ONE Pro in the ``Toy'' dataset, disliked the others.
% Unlike others, ``Casserole'' images are vertically elongated with an aspect ratio of 3:4 to evaluate under different photographing situations. 

\renewcommand{\arraystretch}{1.1}
\begin{table}[b]
  \caption{Quantitative results for \textbf{visible light images} of \textbf{validation data}.}
  \centering
  \setlength{\tabcolsep}{2.0pt}
  \begin{tabular}{c|l|c|c|c}
    \hline\noalign{\smallskip}
       Scene & Method & PSNR ↑ & SSIM ↑ & LPIPS ↓ \\
        \hline
        
              \multirow{4}{*}{Toy} & Ours w/ O.Pro V & 13.80 & 0.55 & 0.64 \\
                                   & BARF w/ O.Pro V & \textbf{20.00} & \textbf{0.75}& \textbf{0.34} \\
                \cline{2-5}
                & Ours w/ iPhone V & \textbf{13.69} & \textbf{0.55} & \textbf{0.56} \\
                & BARF w/ iPhone V & 12.95 & 0.51 & 0.62 \\
              \hline

              \multirow{4}{*}{Casserole} & Ours w/ O.Pro V & 14.72 & 0.50 & 0.57 \\
                                         & BARF w/ O.Pro V &\textbf{ 15.19} & \textbf{0.52} & \textbf{0.55}\\
                  \cline{2-5}
                & Ours w/ iPhone V & \textbf{14.16} & \textbf{0.53} & \textbf{0.52} \\
                & BARF w/ iPhone V & 13.09 & 0.51 & 0.56 \\
              \hline

              \multirow{4}{*}{Trace} & Ours w/ O.Pro V & 15.69 &  0.49 & 0.57 \\
                                         & BARF w/ O.Pro V & \textbf{18.21} & \textbf{0.56} & \textbf{0.49}  \\
                  \cline{2-5}
                & Ours w/ iPhone V & 13.69 & 0.37 & 0.61 \\
                & BARF w/ iPhone V & \textbf{14.26} & \textbf{0.38} & \textbf{0.55} \\
              \hline

    \hline
  \end{tabular}
  \label{tab:result_2-val}
\end{table}
\setlength{\tabcolsep}{1.4pt}

\begin{figure}[t]
  \centering
    \includegraphics[width=0.85\linewidth]{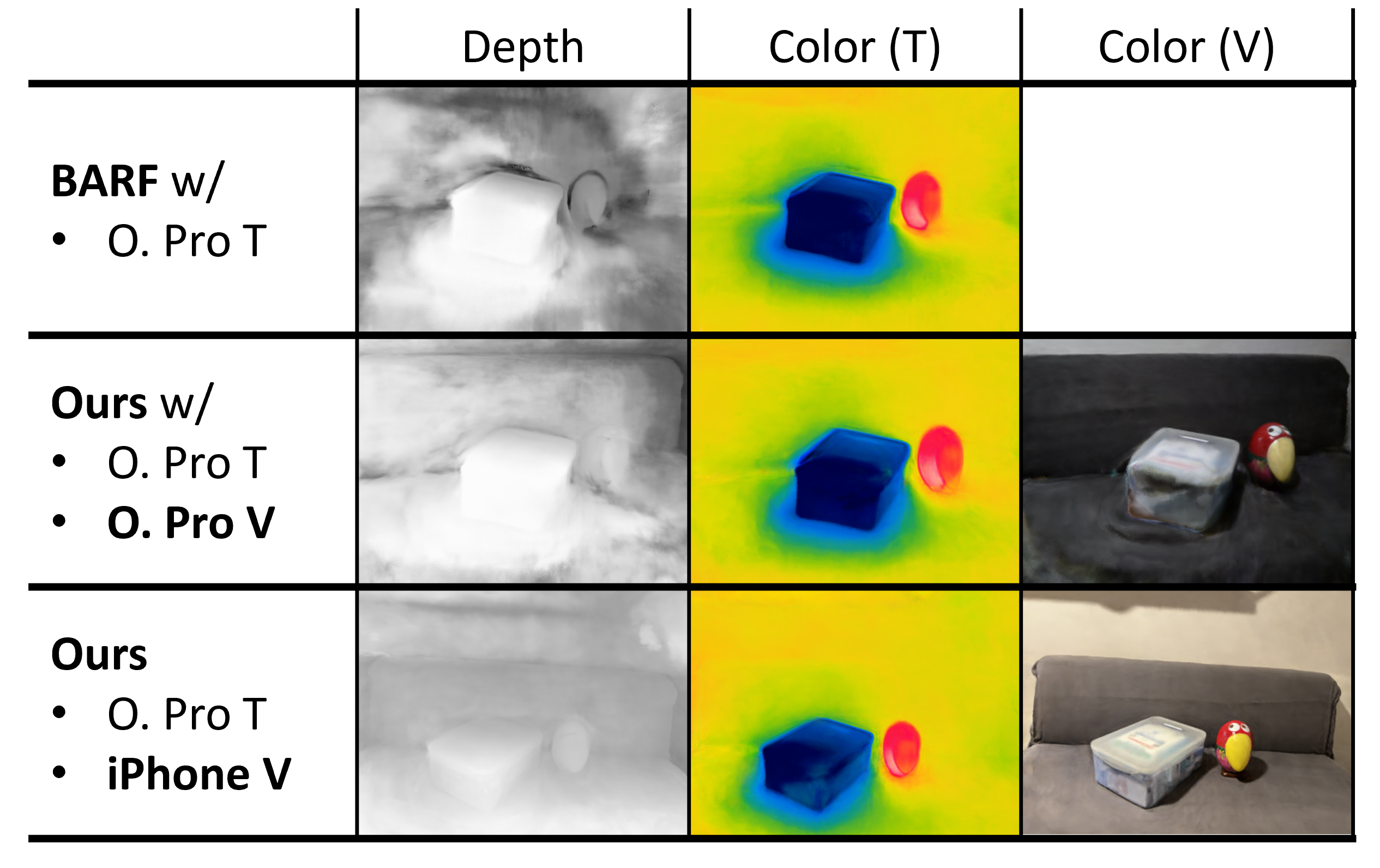}
  \caption{The synthesized results of the ``Toy'' dataset, the practical situation. From the top row, BARF optimized only with thermal images, MultiBARF with FLIR ONE Pro visible, and iPhone visible images.}\label{fig:result_2}
\end{figure}

\leavevmode \\
\textbf{Results.} \cref{fig:result_2} shows the qualitative results. Similar to the experiment for basic performance evaluation, our approach, but only FLIR ONE Pro images, improves density estimation results. However, the combination with iPhone visible light images on the iPhone makes the object shapes clearer than those of FLIR ONE Pro. 

Multiple advantages can improve the quality. First, the camera of the iPhone can capture shadowed areas and detail textures clearer than FLIR ONE Pro, as shown in \cref{fig:datasets} (c) and (d). The iPhone camera has several functions to capture high-quality photos, such as auto image stabilization. The iPhone data contains more images than FLIR ONE Pro because there is no time limit to image shooting. In addition, the angle of view differs between visible light and thermal images. The latter two differences increase overlap among images, improving 3D reconstruction results. These advantages work, too, even in MultiBARF, which uses additional sensor images to learn density. 

Also, quantitative results suggest no systematic degradation over data splits (training/validation) and sensor combinations for synthesizing visible light images (\cref{tab:result_2,tab:result_2-val}). It is even in the ``Toy'' dataset without implicit camera pose relationships.

The results show that Multi-BARF performs well without synchronizing multiple sensors. As shown before, the framerate of the iPhone differs from that of FLIR ONE Pro. That is, the capturing of static scenes with each device is asynchronous. This asynchronization is also an advantage of Multi-BARF. For example, we can use one visible light dataset captured in a good lighting environment and thermal images in a dark situation for static scenes. This advantage makes dataset building easier than having to shoot with close timing.

\subsection{Comparison of optimizing method}\label{sec:exp_train}
We explored the appropriate optimizing method to complete our process. Since sub-MLPs that output color learn only the corresponding subsets of images, the model needs two optimizing modes: optimizing with visible light images and optimizing with thermal images. The model also requires processing to optimize the camera poses as BARF\@. Therefore, we had to build a schedule to make these three processes work appropriately.

\leavevmode \\
\textbf{Datasets.} We show the result on ``Toy''. In this experiment, we compare the optimization method using the combination of visible light images of an iPhone and thermal images of FLIR ONE Pro. Other dataset settings are the same as in the experiment in \cref{sec:exp_practical}.

\leavevmode \\
\textbf{Results.} 
First, we mention the method adopted in this study, which uses the two modes alternatively in this paper. In this case, the model can update modules for a sensor's images depending on the results of another one, which prevents the model from unilateral destructive training with one sensor's images. As a result, this method successfully sorts each sensor texture to post-branch MLPs. Also, this method is consistent with the optimization process of Bundle-Adjusting Neural Radiance Fields. The method starts optimization with low-frequency features due to the camera pose estimation. The alternative method does not interfere with this advantage. 

\begin{figure}[b]
    \centering
      \includegraphics[width=0.90\linewidth]{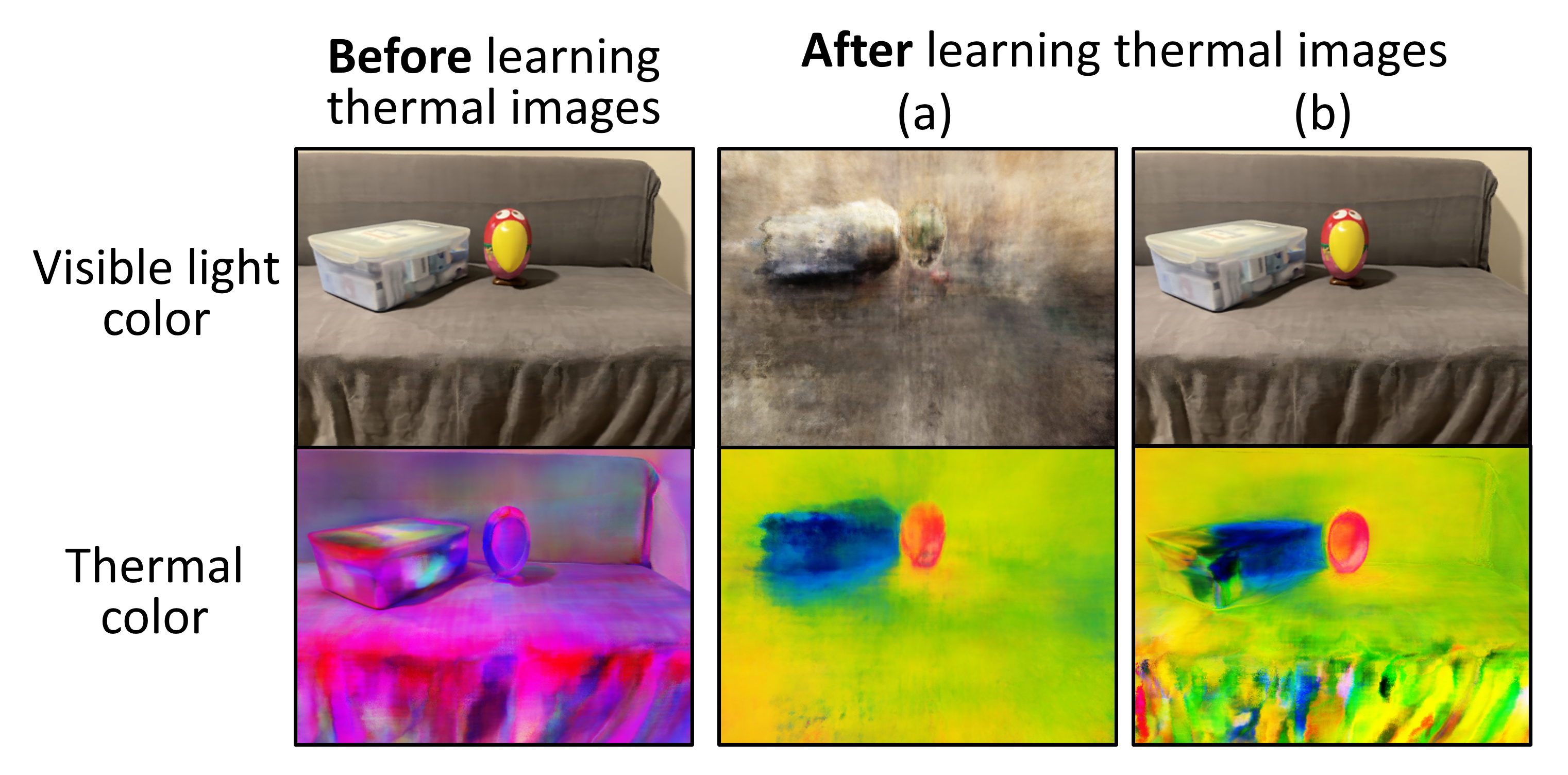}
    \caption{Synthesized images of visible light (upper) and thermal (lower) color channels by the model learned thermal images after completing learning visible light images. The left column shows the results when just finished with visible light images. The remaining two are after learning thermal images. Still, the right is when the part of DNN related to visible light images is frozen during learning thermal images, while the middle is with no frozen.}\label{fig:res_optimmethod}
\end{figure}

Next is a more straightforward method. In this case, the model learns thermal images after completing optimization with visible light images. In this case, we trained our model for 200000 epochs with visible light images and then trained another 200000 epochs with thermal images. As a result, the thermal training phase updated the pre-trained NeRF by visible images destructively, as shown in \cref{fig:res_optimmethod} (a). This result suggests that training with thermal images can update the DNN parameters significantly but cannot generate density distribution appropriately.

The final method also puts the thermal phase after the visible light phase, but the part of the DNN for synthesizing visible light images is frozen. In this case, the synthesized thermal colors changed from the initial condition, but the color boundaries derived from visible images appeared, as shown in \cref{fig:res_optimmethod} (b). This characteristic indicates that the pre-branched MLP recorded the texture feature of visible light, and the thermal color MLP could not repaint it. We can change the thermal color MLP to be thick, but changing the pre- and post-branch layer balance can harm the results. These results suggest that, with this method, it is hard to convert visible light scenes into thermal scenes to change scene appearances by feature embedding, such as Chen et al.~\cite{ha-nerf}. 

As an additional comparison between the adopted and the other two methods, the latter two start optimization with lower-frequency features of thermal images but use pre-constructed density, including high-frequency information, which could partially cause problems in the bundle adjusting process.

%% file: sec/5_conclusion.tex
\section{Conclusion}\label{sec:conclusion}
We introduced MultiBARF, a method replacing the co-registration and geometric calibration by synthesizing pairs of two different sensor images and depth images at assigned viewpoints without complicated camera calibration processes. We evaluated the model and training method on visible light and thermal image datasets. Through the experiments, we confirmed that our approach could generate NeRF consisting of two sensor colors and register the sensor color distributions.

The next step is quantitatively evaluating registration and 3D reconstruction accuracy and improving these accuracies. Also, our future work involves developing a method to express transparent objects that are hard to see in visible light images and LiDAR but appear in the thermal infrared region. General transparent materials such as glass and acrylic plates appear clear in thermographic images but not in visible light images. We will work to utilize multi-sensor combinations that are easily applicable to more everyday situations.